\documentclass[twoside,11pt]{article}

\usepackage{blindtext}

\usepackage{jmlr2e}

\usepackage[utf8]{inputenc} % allow utf-8 input
\usepackage[T1]{fontenc}    % use 8-bit T1 fonts
\usepackage{hyperref}       % hyperlinks
\usepackage{url}            % simple URL typesetting
\usepackage{booktabs}       % professional-quality tables
\usepackage{amsfonts}       % blackboard math symbols
\usepackage{nicefrac}       % compact symbols for 1/2, etc.
\usepackage{microtype}      % microtypography
\usepackage{xcolor}         % colors
\usepackage{amsmath}
\usepackage{bm}
\usepackage{pgfplots}
\usepackage{tikz}
\usepackage{subcaption}
\usepackage{wrapfig}
\usepackage{marvosym}
\pgfplotsset{compat=1.17} 
\usepackage{color,colortbl}
\definecolor{cycle2}{RGB}{55, 126, 184}
\usepackage[capitalise,noabbrev]{cleveref}

\def\x{{\pmb x}}

\def\X{{\pmb X}}
\def\A{{\pmb A}}

\def\Z{{\pmb Z}}
\def\H{{\pmb H}}
\def\h{{\pmb h}}
\def\G{{\mathcal G}}
\def\R{{\mathbb R}}
\def\T{{\pmb T}}

\def\S{{\mathcal S}}

\usepackage{lastpage}
\jmlrheading{25}{2024}{1-\pageref{LastPage}}{5/23; Revised
8/23}{3/24}{23-0680}{Alexandre Duval, Victor Schmidt, Santiago Miret, Yoshua Bengio, Alex Hernandez Garcia, David Rolnick}

\ShortHeadings{PhAST}{Duval, Schmidt et al.}
\firstpageno{1}

\begin{document}

\title{PhAST: Physics-Aware, Scalable, and Task-Specific GNNs for Accelerated Catalyst Design}

% TODO: figure out how to do "equal contributions" properly
\author{%
  \name Alexandre Duval \email{alexandre.duval@mila.quebec} \\
  \addr Mila, Inria, CentraleSupelec\\
  \AND
  \name Victor Schmidt \email{schmidtv@mila.quebec} \\
  \addr Mila, Universtité de Montréal \\
  \AND
  \name Santiago Miret \email{santiago.miret@intel.com} \\
  \addr Intel Labs\\
  \AND
  \name Yoshua Bengio \email{yoshua.bengio@mila.quebec} \\
  \addr Mila, Université de Montréal, CIFAR Fellow\\
  \AND
  \name Alex Hernández-García \email{alex.hernandez-garcia@mila.quebec} \\
  \addr Mila, Université de Montréal\\
  \AND
  \name David Rolnick \email{david.rolnick@mila.quebec} \\
  \addr Mila, McGill University\\
}

\editor{Shakir Mohamed}

\maketitle

\begin{abstract}
\end{abstract}%   <- trailing '%' for
Mitigating the climate crisis requires a rapid transition towards lower-carbon energy. Catalyst materials play a crucial role in the electrochemical reactions involved in numerous industrial processes key to this transition, such as renewable energy storage and electrofuel synthesis. To reduce the energy spent on such activities, we must quickly discover more efficient catalysts to drive electrochemical reactions. Machine learning (ML) holds the potential to efficiently model materials properties from large amounts of data, accelerating electrocatalyst design. The Open Catalyst Project OC20 dataset was constructed to that end. However, ML models trained on OC20 are still neither scalable nor accurate enough for practical applications. In this paper, we propose task-specific innovations applicable to most architectures, enhancing both computational efficiency and accuracy. This includes improvements in (1) the graph creation step, (2) atom representations, (3) the energy prediction head, and (4) the force prediction head. We describe these contributions, referred to as PhAST, and evaluate them thoroughly on multiple architectures. Overall, PhAST improves energy MAE by 4 to 42$\%$ while dividing compute time by 3 to 8$\times$ depending on the targeted task/model. PhAST also enables CPU training, leading to 40$\times$ speedups in highly parallelized settings. Python package: \url{https://phast.readthedocs.io}.


\section{Introduction}
\label{sec:intro}

To mitigate climate change at a global scale, it is imperative to reduce the carbon emissions of ubiquitous industrial processes like cement production or fertiliser synthesis, as well as to develop infrastructures for storing low-carbon energy at scale, enabling to re-use it wherever and whenever needed. Since such processes rely on electrochemical reactions, they require the design of more efficient electrocatalysts \citep{zakeri2015hes} to become more environmentally and economically viable.

However, discovering easy-to-exploit low-cost catalysts that drive electrochemical reactions at high rates remains an open challenge. In fact, today's catalyst discovery pipeline mostly relies on expensive quantum mechanical simulations such as the Density Functional Theory (DFT) to approximate the behaviour of the materials involved in the targeted chemical reaction. Unfortunately, the high computational cost of these simulations limits the number of candidates that can be efficiently tested, and consequently stagnates further advances in the field. 

Machine learning (ML) holds the potential to approximate these calculations while reducing the time needed to assess each candidate by several orders of magnitude \citep{zitnick2020introduction}. This capability could transform the search for new catalysts, by making it possible to sort through millions or even billions of possible materials to identify promising candidates for experimental inquiry\citep{zitnick2020introduction}. 

To enable to use of ML for catalyst discovery, the Open Catalyst Project released OC20 \citep{chanussot2021open}, a large data set of pairs of catalyst and target molecule (known as \textit{adsorbate}), along with the \textit{relaxed energy} of the resulting system---a relevant metric to assess how good a catalyst is for a given chemical reaction---computed with DFT from the initial atomic structure. Despite recent progress~\citep{gasteiger2021gemnet, ying2021transformers}, major challenges remain. First, state-of-the-art models have not yet reached high enough performance for practical applications. Second, they are still too computationally expensive to allow the millions of inferences required to explore the large space of potential catalysts. Third, the graph neural networks (GNNs) typically used are designed for general 3D material modeling tasks rather than specifically for catalyst discovery, a complex task that may benefit from domain-specific architectures.

To address these challenges, we propose multiple model improvements to increase the accuracy and scalability of generic GNNs applied to catalyst discovery. In particular, our contributions are (1) a graph construction that is tailored to catalyst-adsorbate modeling, (2) richer physics-based atom representations, (3) an energy head that learns a weighted sum of per-atom predictions, and (4) a direct force prediction head encouraging energy conservation. We provide a broad evaluation of these contributions on OC20 and a thorough ablation study. In sum, the proposed PhAST improvements decrease energy MAE by 5--42~\% while dividing compute time by 3--8$\times$ depending on the targeted task/model. These gains in model scalability enable efficient CPU training, with up to 40$\times$ speedups in highly parallelized pipelines using PhAST, making these models significantly more accessible to a wider community of researchers. We also believe that our work provides valuable insights for future research as it leverages domain-specific knowledge to improve parts of the pipeline that were not investigated up to now. Overall, the resulting performance and scalability gains open the door to a practical use of GNNs for new electrocatalyst design, the ultimate end goal of this line of research.

%%%%%%%%%%%%%%%%%%%%%%%%%%%%%%%%%%%%%%%%
% BACKGROUND
%%%%%%%%%%%%%%%%%%%%%%%%%%%%%%%%%%%%%%%%

\section{Background}
\label{sec:rw}

\begin{figure}[h]
    \centering
    \includegraphics[width=\textwidth]{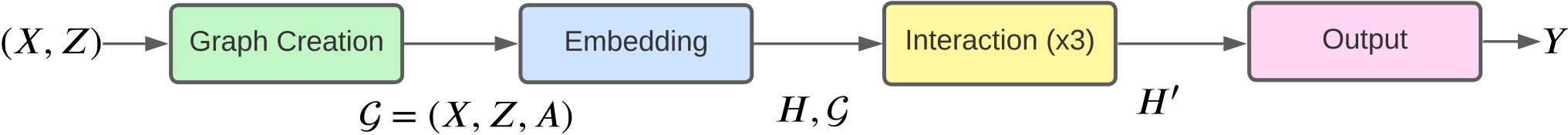}
    \caption{Common GNN inference pipeline for 3D material modeling. The graph creation step remains unchanged across all methods: it creates $\A$ using cutoff distances and periodic boundary conditions. The Embedding and Output blocks slightly differ across models but the underlying idea is the same. The Embedding block learns a representation for each chemical element and the Output block applies a global pooling of each node's representation to obtain the energy prediction. The key distinction between methods typically lies in the Interaction block, where the message passing schemes vary.}
    \label{fig:pipeline}
\end{figure}

The problem we address is the prediction of the relaxed energy $y \in \R$ of an adsorbate-catalyst system from its initial configuration in space $(\X, \Z)$, where $\X \in \R^{N \times 3}$ is the matrix of 3D atom positions and $\Z \in \mathbb{N}^{N}$ contains atom characteristic numbers. This task is referred to as \textit{Initial Structure to Relaxed Energy}, IS2RE, in \cite{zitnick2020introduction}.  This is commonly formulated as a graph regression task, where each sample is represented as a 3D graph $\G$ with node set $\mathcal{V}$ of dimension $N$ and adjacency matrix $\A \in \R^{N \times N}$. $\H \in \R^{N \times H}$ represents atom embeddings
and $\T \in \{0, 1, 2\}^{N}$ corresponds to tag information (see \ref{sec:graph-creation}). 
ML models designed for this task generally adopt graph neural networks as an architecture, as it naturally suits 3D material modeling. Such GNNs typically share a common pipeline for how they are applied, as depicted in Fig.~\ref{fig:pipeline}. 

In material modeling tasks, it is desirable to endow ML models with relevant symmetry properties. In particular, we want predictions to be equivariant to translations, rotations and (often) reflections. Many models enforce these physical priors within the architecture, making it explicitly invariant or equivariant to the desired transformations. Formal definitions are included in \cref{app:symmetries}. 

Many GNNs in prior work focus on enforcing equivariance, though it is not strictly required for relaxed energy prediction, which calls for invariance. Equivariant GNNs \citep{thomas2018tensor, anderson2019cormorant, fuchs2020se, batzner20223, brandstetter2021geometric} are expressive and generalize well, but are very computationally expensive as they are constrained by equivariant filters built on spherical harmonics and the Clebsch-Gordan tensor product. Recent methods ~\citep{schutt2021equivariant, satorras2021n, tholke2022torchmd} model equivariant interactions in Cartesian space using both invariant (scalar) and vector representations. While they are faster, their architectures are often very complex and lack theoretical guarantees. 
Alternatively, E(3)-invariant methods \citep{schutt2017schnet, unke2019physnet, shuaibi2021rotation, ying2021transformers, adams2021learning, zitnick2022spherical} do not use atom positions directly in their internal workings. Instead, these methods extract and use quantities that remain invariant under rotations and reflections. 
DimeNet++ \citep{klicpera2020directional, klicpera2020fast}, for example, includes a directional message passing (MP) mechanism that incorporates bond angles in addition to atom relative distances. However, distances and bond angles do not suffice to uniquely identify the graph 3D structure. This is achieved by SphereNet \citep{liu2021spherical} and GemNet \citep{gasteiger2021gemnet}, which additionally extract torsion information (between quadruplets of nodes). On the downside, these methods are very computationally expensive as they require considering 3-hop neighbourhoods for each update step. Importantly, all these Message Passing methods aim at broad applicability and do not leverage the specific constraints of individual tasks.

%%%%%%%%%%%%%%%%%%%%%%%%%%%%%%%%%%%%%%%%
% METHOD
%%%%%%%%%%%%%%%%%%%%%%%%%%%%%%%%%%%%%%%%

\section{Proposed Method}
\label{sec:method}

In this section, we describe PhAST, a Physics-Aware, Scalable, and Task-specific GNN framework for catalyst design. Notably, the architectural innovations in our proposed framework are applicable to most current GNNs used in materials discovery. These include a novel graph creation step, richer atom representations, an advanced energy head for graph-level prediction, and a direct (energy-conserving) force-head for atom-wise predictions.

\subsection{Graph creation}
\label{sec:graph-creation}

Although the graph construction step is critical in graph ML tasks, it has received little or no attention by previous work on the OC20 data set. Most methods reuse the original proposal by \citet{chanussot2021open}. In OC20, each graph's atom is tagged as part of the adsorbate (tag 2), the catalyst's surface (tag 1), or its sub-surface volume (tag 0). Tag 0 atoms were originally added in DFT simulations to represent more explicitly the repeating pattern of the catalyst slab. They are, by definition, further away from the adsorbate and are fixed by construction, unlike tag 1 and tag 2 atoms, which can move during the relaxation. As a result, we hypothesise they contain redundant information, making them of lesser importance to predict the final relaxed energy. Besides, since they account for $\sim 65\%$ of the nodes~\ref{app:node-count}, and since SOTA GNNs often depend on multiple hops to compute bond and torsion angles, we propose to remove these nodes from the graph. This should greatly reduce inference time without impacting expressivity. As an alternative intermediate approach, we explore forming \textit{super nodes} that aggregate tag 0 atoms to avoid a potential information loss caused by their total removal. We briefly describe these changes below, with more details in \cref{app:sec:graph-creation}.

\textbf{remove-tag-0} removes all atoms with tag 0 (i.e. in $\mathcal{S} = \{i \in \mathcal{V}: t_i=0\}$) from the graph, adapting correspondingly all graph attributes ($\X$, $\A$, $\Z$, $\T$, etc.).
    
\textbf{one-supernode-per-graph} aggregates all tag 0 nodes from $\G$ into a supernode $s$ with position $\x_s = \frac{1}{|\S|} \sum_{i \in \S} \x_i$, adjacency $A_{is} = \max(A_{ij}: j \in \S)$ and a new characteristic number $z_s$.
% z_s = \{z_i: i \in \S\}
    
\textbf{one-supernode-per-atom-type} replicates the above strategy but creates one super node per distinct chemical element in the catalyst subsurface. Its attributes are defined based on its components, as previously.

% -------------------------------------------------------------------------------------------

\subsection{Atom Embeddings}
\label{sec:atom-embed}

In all previously proposed GNN methods to solve energy-prediction tasks (e.g. IS2RE), atom representations are learned from scratch based on atomic number $\H = \H_{\Z}$. We propose to leverage domain information to improve these representations. 
First, we hypothesise that whether a given atom belongs to the adsorbate, the catalyst surface, or its subsurface is important information. We therefore incorporate tag information into our model by utilizing a learnable embedding matrix $\H_{\T}$ that encodes the tags as vectors. Second, we know from previous studies that some atomic properties (e.g.~atomic radius or density) are useful for catalyst discovery \citep{takigawa2016machine, ward2017including}. We leverage them as an additional embedding vector $\H_{\mathbf{F}}$ (see \ref{app:sec:sn-type} for the full list of properties).
Lastly, we let the GNN learn embeddings for both the group and period information ($\H_{\mathbf{P}}, \H_{\mathbf{G}}$) since atoms belonging to the same group or period often share similar behaviours \citep{xie2018crystal}. As a result, our proposed atom embedding $\H$ is a concatenation of all of the above: $\H = \H_Z || \H_T || \H_F || \H_P || \H_G$.

% -------------------------------------------------------------------------------------------

\subsection{Energy head}
\label{sec:energy-head}

In the literature, there is often limited focus  on the energy head, which is the part of the output block responsible for the energy computation from final atom representations $\h^L_{i}$. To the best of our knowledge, all GNNs compute the relaxed energy using global pooling $\hat{y} = \sum_{i \in \mathcal{V}} h_{i}$, where node embeddings are reduced to a scalar $h_{i}$ by linear layers. We identify two limitations in this procedure: First, all atoms are assigned the same importance, even though the properties of an atom are normally influenced by the properties of the element. Second, the graph topology is neglected by simply summing all atom encodings regardless of their 3D positions. To overcome these limitations, we explore alternative energy heads.

First, \textbf{a weighted sum of node representations}, which grants adaptive importance to each chemical element, expressed as $\hat{y} = \sum_{i \in \mathcal{V}} \alpha(\h^{L}_i) \cdot h_{i}$ or $\hat{y} = \sum_{i \in \mathcal{V}} \alpha(\h^{0}_i) \cdot h_{i}$, where the learnable importance weights $\alpha(\cdot)$ depend either on the embedding block initial encodings $\h^{0}_{i}$ or final ones $\h^{L}_{i}$. 

Second, \textbf{a hierarchical pooling approach} endowed with the following energy head pipeline: $\h^{L}_{i} \rightarrow $ [Pooling $\rightarrow$ GCN] ($\times 2$) $\rightarrow$ Global Pooling $\rightarrow$ MLP $\rightarrow \hat{y}$. By applying a graph convolutional network \citep[GCN,][]{kipf2016semi} on a coarsened graph, we propagate information differently, allowing us to capture hierarchical graph information. We implement \textsc{hoscpool} \citep{duval2022higher}: an end-to-end pooling operator that learns a cluster assignment matrix using a loss function inspired by motif spectral clustering.% Contrary to \textsc{graclus}, it leverages node features, captures higher-order connectivity patterns and is differentiable. 

\subsection{Force head}
\label{sec:force-head}

A closely related task to IS2RE (i.e. energy prediction) consists in computing \textit{forces} together with energy. This involves the additional prediction and training of atom-wise 3D vectors representing the forces currently applied on each atom by the rest of the system. This task is referred to as \textit{Structure to Energy and Forces}, S2EF in \cite{zitnick2020introduction}. In many previous works, atomic forces are directly computed as the predicted energy's gradient with respect to atom positions $\vec{y}_i = \text{-} \frac{\partial y}{\partial \x_i}$ (i.e. its definition in physics). While this guarantees energy-conserving forces\footnote{a desirable feature in molecular dynamics, as it improves the stability of the simulation and the ability to reach local minima \citep{chmiela2017machine}}, \citet{kolluru2022open} demonstrated the significant computational burden associated with this approach, which increases memory use by a factor of 2–4 and leads to decreased modeling performance for specific datasets.
As a result, several recent works neglected this principle on OC20, proposing \textbf{direct force predictions} from final atom representations. 
Here, we extend this idea by proposing a plug-and-play force head architecture for traditional energy conserving GNNs \citep{schutt2017schnet, klicpera2020directional}. We thus use a shared backbone with two independent output heads: $\Phi^E$ denotes graph-level energy predictions and $\Phi^F$ denotes atom-level force predictions (both including the backbone). 

We keep the traditional \textbf{energy loss} $\mathcal{L}_E = ||\hat{y} - y||_2$ and \textbf{force loss} $\mathcal{L}_F = \sum_i ||\hat{\vec{y}}_i - \Vec{y}_i||_2$ to train our network, respectively pushing predicted energy $\hat{y} = \Phi^E(\mathcal{G})$ towards its ground truth value $y$ and predicted forces $\hat{\vec{y}}_i= \Phi_i^F(\mathcal{G})$ towards their ground truth value $\vec{y}_i$, i.e. the negative energy gradient. However, using $\mathcal{L}_E$ and $\mathcal{L}_F$ does not guarantee that $\hat{\vec{y}}_i$ and $\text{-} \frac{\partial \hat{y}}{\partial \x_i}$ will be aligned since both are predicted separately.

To encourage energy conservation in the presence of a force head, we propose a new \textbf{gradient-target loss} term: the $L_2$ (squared) distance between atomic force predictions and the negative energy gradient with respect to atom positions. For a given graph:
\begin{align}
    % \mathcal{L}_{EC} &= \sum_i \big\Vert \Phi^F(\mathcal{G}) - (- \nabla \Phi^E)(\mathcal{G}) \Vert_2^2 \\
    %\mathcal{L}_{EC} &= \big\Vert \Phi^F(\mathcal{G}) - (- \nabla \Phi^E)(\mathcal{G}) \big\Vert_2^2  \nonumber \\
    \mathcal{L}_{Grad} &= \frac{1}{|G|} \sum_{i\in\mathcal{G}} \big\Vert \hat{\vec{y}}_i - \big(\text{-}\frac{\partial \hat{y}}{\partial \x_i}\big) \big\Vert_2^2
    \label{eq:energy-conserving}
\end{align}

This term aims to correct for the possible misalignment between predicted forces and predicted energy, reinforcing the energy conserving character of our predictions. Alternatively, while the norm considers both the concept of direction and distance, we also study substituting the above term by a \textbf{cosine similarity loss} between predicted forces and the negative predicted energy gradient which mostly focuses on directional misalignment:

\begin{equation}
    % \mathcal{L}_{EC} = cos\_sim(\hat{\vec{y}}, \vec{y})
    \label{eq:cosine-similarity}
    \mathcal{L}_{Cos} = \frac{1}{|G|} \sum_{i\in\mathcal{G}} \dfrac{ \hat{\vec{y}}_i \cdot \big(\text{-}\frac{\partial \hat{y}}{\partial \x_i}\big)}{\Vert \hat{\vec{y}}_i \Vert _2 \cdot \Vert \big(\text{-}\frac{\partial \hat{y}}{\partial \x_i}\big) \Vert _2}.
\end{equation} 

While one could train with $\mathcal{L}_{Grad}$ or $\mathcal{L}_{Cos}$ all along, we empirically obtained slightly better performance by using this term at the end of training only, as fine-tuning. We hypothesize this is because we first want the GNN to properly predict energy and forces before making predictions more energy conserving.

\subsection{PhAST: final components}
\label{subsec:final-model}
In~\Cref{sec:ablation}, we present the ablation study that lead us to the selection of the various components that make PhAST, across the four areas of improvements. All results displayed in \cref{sec:eval} thus leverage these components.
As an overview, we list them here:
\begin{enumerate}
    \item Graph creation: \textit{remove-tag-0}.
    \item Atom embeddings: all embeddings $\H = \H_Z || \H_T || \H_F || \H_P || \H_G$.
    \item Energy head: weighted sum from initial embeddings.
    \item Force head: direct force prediction with gradient target loss.
\end{enumerate}

%%%%%%%%%%%%%%%%%%%%%%%%%%%%%%%%%%%%%%%%
% EVALUATION
%%%%%%%%%%%%%%%%%%%%%%%%%%%%%%%%%%%%%%%%

\section{Evaluation}
\label{sec:eval}
In this section, we evaluate the performance and scalability of the PhAST framework for five well-known GNNs on the OC20 dataset \citep{chanussot2021open}. We first perform an in-depth study of the first three components of PhAST on the OC20 IS2RE energy prediction task, before looking at the gains of the last component (i.e.~the force head) on the OC20 S2EF-2M energy-force prediction task.

\subsection{Baselines}
\label{sec:ablation-baselines}

We target five well-known GNN baselines to study the impact of our contributions, including state-of-the-art method GemNet-OC \cite{gasteiger2022gemnet}. We have selected them based on their popularity and ease-of-implementation but note that PhAST improvements are applicable to all recent GNNs for 3D material modeling, to the best of our knowledge, because the changes are architecture-agnostic. We use the hyperparameters, training settings and model architectures provided in the original papers. As mentioned in \cref{fig:pipeline}, they all follow a similar pipeline, mainly differing in their interaction blocks, which we briefly detail below.

\textbf{SchNet} \citep{schutt2017schnet} is a simple message passing architecture that leverages relative distances to update atom representations via a continuous filter: $\h_i^{(l+1)} = \sum_j \h_j^l \odot W^l(\x_i-\x_j)$ where $W^l(\x_i-\x_j)$ is a radial basis function to encode distance between atom pairs. 

\textbf{DimeNet++} \citep{klicpera2020fast} is an optimised version of DimeNet \citep{klicpera2020directional}, which proposes a directional message passing. In other words, they compute and update edge representations instead of atoms) using interatomic distances $\mathbf{e}_{RBF}$ (encoded via bessel functions) and bond angles $\mathbf{a}_{SBF}$ (encoded via 2D spherical Fourier-Bessel basis): 
\begin{equation*}
    \mathbf{m}_{ij}^{(l+1)} = f_{update}\big( \mathbf{m}_{ij}^{(l)}, \sum_{k \in N_j \setminus i} f_{int}(\mathbf{m}_{ij}^{(l)}, \mathbf{e}_{RBF}^{(ij)}, \mathbf{a}_{SBF}^{(ki,ji)}) \big)
\end{equation*}

\textbf{ForceNet} \citep{hu2021forcenet} is a scalable force-centric GNN that does not impose explicit physical constraints (energy conservation, rotational invariance). It attempts to encourage invariance by efficient rotation-based data augmentation. Model-wise, it adopts a node message passing approach that leverages node positions directly via a spherical harmonics basis.

\textbf{GemNet} \citep{gasteiger2021gemnet} builds on top of DimeNet++, but additionally incorporates torsion information between quadruplets of atoms. This grants it the ability to process more geometric information and thus to distinguish between a wider range of different graphs, at the cost of extra computational cost and model complexity.

\textbf{GemNet-OC} \citep{gasteiger2022gemnet} is an improved version of GemNet.

\subsection{PhAST performance on IS2RE}
\label{subsec:is2re}

\textbf{Dataset}. OC20 contains 1,281,040 DFT relaxations of randomly selected catalysts and adsorbates from a set of plausible candidates. In this section, we focus on the \textit{Initial Structure to Relaxed Energy} (IS2RE) task \citep{zitnick2020introduction}, that is the direct prediction of the relaxed adsorption energy from the initial atomic structure. It comes with a pre-defined train/val/test split, 450,000 training samples and hidden test labels. Experiments are evaluated on the validation set, which has four splits of $\sim 25K$ samples: In Domain (ID), Out of Domain adsorbates (OOD-ads), Out of Domain catalysts (OOD-cat), and Out of Domain adsorbates and catalysts (OOD-both). 

\begin{figure}[t]
    \centering
    \includegraphics[width=\textwidth]{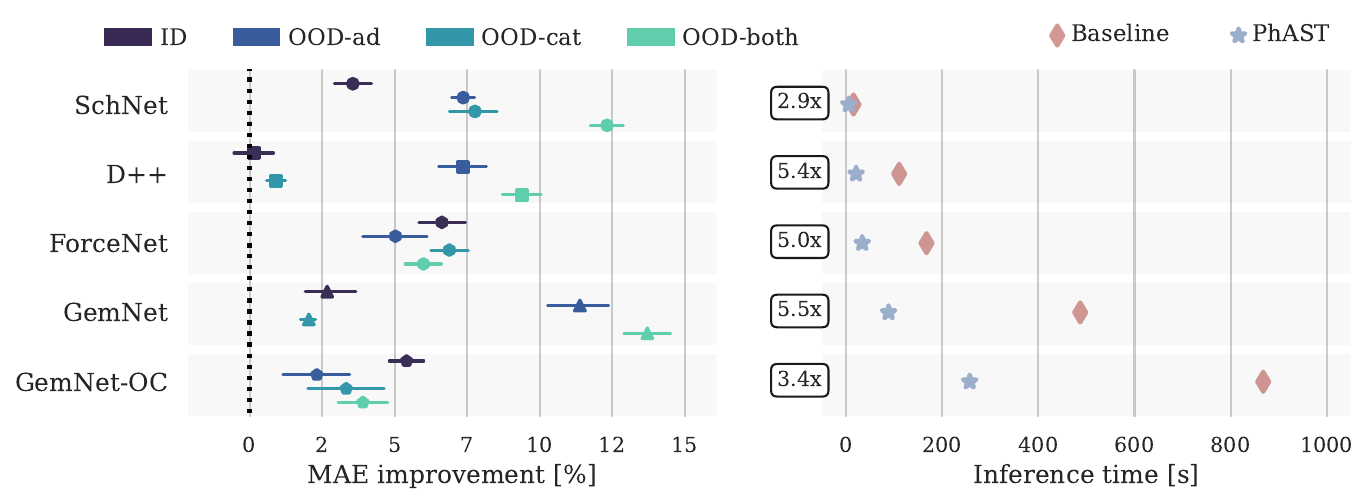}
    \caption{A comparison of the improvements brought by PhAST to the model's MAE (left) and inference times (right) on \textsl{OC20} IS2RE. The PhAST components are selected in \cref{sec:ablation} and summarised in \cref{subsec:final-model}. MAE improvements (left) are computed as in Equation~\ref{eq:mae_improvement} and values to the right of the dashed line at 0.0 denote an improvement of PhAST with respect to the baseline. The results are averaged over 3 runs, with bootstrapped confidence intervals represented by small horizontal bars. PhAST leads to a significant MAE improvement for each validation split, up to $13~\%$, in addition to decreasing inference time (s) by several factors (e.g. 3.4$\times$ or 5.4$\times$). Absolute numerical values are provided in \cref{tab:best-perf-is2re}.}
    \label{fig:summary}
\end{figure}

\textbf{Metrics}. We measure accuracy via the energy \textit{Mean Average Error} (MAE) in meV on each validation split, and scalability by the \textit{inference time} (in seconds) over the whole ID validation set. We also include \textit{throughput} in \cref{app:tab:throughput-is2re}, i.e. the number of samples processed per second at inference time\footnote{\textit{Throughput} differs from \textit{inference time} as it only measures the on-device forward pass of the model, neglecting data-loading, inter-device transfers etc. While more theoretically relevant, it is also less practically informative, which is why we report both.}. Since the absolute time metrics are difficult to compare across hardware setups with respect to other works, we note that the most relevant metrics are the \textit{relative} improvements we show using the exact same hardware and software for all models. 
In order to easily visualize the contributions of PhAST on the baseline models in terms of performance improvement, in Figure~\ref{fig:summary} (left) we plot the relative MAE improvement with respect to the baseline. Specifically, we compute the MAE improvement as 
\begin{equation}
\text{MAE improvement} = 100 \times \frac{\text{MAE(baseline)} - \text{MAE(PhAST)}}{\text{MAE(baseline)}}.
\label{eq:mae_improvement}
\end{equation}

\textbf{Baselines}. We study the enhancements brought by the PhAST components (see \cref{subsec:final-model}) to five key GNN architectures for material modeling: SchNet \citep{schutt2017schnet}, DimeNet++ \citep{klicpera2020fast}, ForceNet \citep{hu2021forcenet}, GemNet \citep{gasteiger2021gemnet} and GemNet-OC \citep{gasteiger2022gemnet}. We compare every baseline with their PhAST counterpart, incorporating the best components of each category detailed in~\Cref{subsec:final-model}, that is graph creation (\ref{sec:graph-creation}), enriched atom embedding (\ref{sec:atom-embed}) and advanced energy-head (\ref{sec:energy-head}), as determined by the ablation study conducted in \cref{sec:ablation}.

\begin{table*}[t]
\centering
\resizebox{\textwidth}{!}{\begin{tabular}{lccccc|c}
\cmidrule[1.3pt]{1-7}
\textbf{Baseline / MAE} & ID & OOD-ad & OOD-cat & OOD-both & Average & Inference time (s) \\
\cmidrule[.5pt]{1-7}
\textsl{SchNet}     & $637$ & $734$ & $661$ & $703$ & $683$ & $15 \pm 0.49$ \\
\rowcolor{cycle2!8} \textsl{PhAST-SchNet}    & $618$ & $677$ & $611$ & $616$ & $\textbf{630}$  & $\pmb{5 \pm 0.36}$\\
\cmidrule[.5pt]{1-7}
\textsl{D++}    & $571$ & $722$ & $561$ & $661$ & $628$  & $110 \pm 0.57$ \\
\rowcolor{cycle2!8} \textsl{PhAST-D++}    & $568$ & $654$ & $560$ & $597$ & $\textbf{595}$  & $\pmb{20 \pm 0.63}$ \\
\cmidrule[.5pt]{1-7}
\textsl{ForceNet}    & $658$ & $701$ & $632$ & $628$ & $654$  & $167 \pm 0.96$ \\
\rowcolor{cycle2!8} \textsl{PhAST-ForceNet}   & $612$ & $664$ & $592$ &  $597$ & $\textbf{616}$  & $\pmb{33 \pm 0.60}$ \\
\cmidrule[.5pt]{1-7}
\textsl{GemNet}    & $573$ & $808$ & $571$ & $744$ & $674$  & $487 \pm 0.24$ \\
\rowcolor{cycle2!8} \textsl{PhAST-GemNet}   & $559$ & $713$ & $558$ & $648$ & $\pmb{619}$  & $\pmb{88 \pm 0.10}$ \\
\cmidrule[.5pt]{1-7}
\textsl{GemNet-OC}    & $593$ & $658$ & $605$ & $584$ & $610$ & $868 \pm 1.89$ \\
\rowcolor{cycle2!8} \textsl{PhAST-GemNet-OC}   & $564$ & $636$ & $587$ &  $562$ & $\textbf{588}$  & $\pmb{257 \pm 0.09}$ \\
\cmidrule[1.3pt]{1-7}
\end{tabular}}
\caption{MAE and inference time for various GNNs and their PhAST counterpart on \textsl{OC20} IS2RE, averaged over 3 runs. \textit{Average} MAE is computed over all validation splits. PhAST models all show improved accuracy and drastic speedups. Note that PhAST-SchNet almost matches the original DimeNet++ while being 21 times faster. A graphical visualisation of the inference time and of relative MAE improvement is provided in Figure~\ref{fig:summary}. Seamingly suboptimal MAE results for the two baseline GemNet models are explained in \Cref{app:gemnets}.} 
\label{tab:best-perf-is2re}
\end{table*}

\textbf{Results}. From \cref{tab:best-perf-is2re} and \cref{fig:summary}, we conclude that our set of PhAST enhancements consistently improve both MAE and inference time upon the original baselines. More precisely, PhAST improves \textit{Average} MAE over the four validation splits by $\sim 6.2~\%$ on average across baselines, while reducing model inference time by $\sim 4.5\times$ (on average across all baselines). Moreover, we observe an MAE improvement of 12.4~\% for SchNet and 9.7~\% for DimeNet++ on val OOD-both, compared to a 7.7~\% and 5.2~\% for \textit{Average} MAE. This suggests that PhAST models generalise better than original baselines. From the ablation study conducted in \cref{sec:ablation}, we conclude that this is due to the combination of our extensions, as they all contribute to significantly better performance on out-of-distribution adsorbate-catalyst systems (OOD-both). Note that inference time gains with PhAST are almost doubled from SchNet, a 1-hop message passing (MP) approach, to DimeNet++, a 2-hops MP approach (from 3$\times$ to 5.5$\times$ speedup)\footnote{A ``limited'' speedup of 3.4$\times$ on GemNet-OC can be explained by the fact that it is a bigger but more efficient version of the original GemNet architecture.}. Throughput scores provided in \cref{app:tab:throughput-is2re} support the results obtained from inference time.

%%%%%%%%%
% S2EF
%%%%%%%%%

\subsection{PhAST performance on S2EF}
\label{subsec:s2ef}

\textbf{Dataset}. In this section, we focus on the \textit{Structure to Energy and Forces} (S2EF-2M) OC20 dataset, that is, the prediction of both the overall energy and atom forces, from a set of 2 million 3D material structures. According to the dataset creators, the 2M split closely approximates the much more expensive full S2EF dataset, making it suitable for model evaluation~\citep{gasteiger2022gemnet}. It also come with pre-defined train/val/test splits\footnote{Similarly to IS2RE, the S2EF validation dataset comes in 4 distinct splits with a cumulative total of 1M samples: ID, OOD-ad, OOD-cat, OOD-both.}.

\textbf{Metrics}. Both Energy MAE (E-MAE) and Forces MAE (F-MAE) are used to measure model accuracy. Regarding scalability, we continue to use the inference time (seconds) over the ID validation set as well as the number of samples per seconds processed by the model at inference time (throughput). 

\textbf{Baselines}. We re-use the same baselines as above, leaving aside GemNet and GemNet-OC given the increased computational scale of this new dataset and the size of those two models\footnote{For reference, GemNet-OC is trained for 2800 GPU hours, a computational budget we could not afford.}. Since ForceNet already has a direct force prediction head, unlike SchNet and D++, we implemented ForceNet-FE which computes forces as the gradient of the energy with respect to atom positions (denoted \textit{FE}, i.e. from energy) in order to assess the added value of the force head. 
\textit{PhAST-FE} includes the components of the previous subsection (graph creation, atom embedding, energy head) and computes forces using the energy gradient while PhAST additionally contains the best performing force head, determined in \cref{sec:ablation}.

\textbf{Results}. From \cref{tab:best-perf-s2ef}, we conclude that (1) \textit{PhAST-FE} improvements are also very significant on S2EF. They lead to better modeling accuracy ($13\%$ E-MAE improvement) and lower compute time (inference time divided by $4.4$) across all three models. (2) Including the proposed PhAST force-head yields significantly better energy MAE than original \textit{PhAST-FE} and it reduces memory usage by a factor of 2 to 4 as well as compute time by a smaller factor. 

PhAST improves Energy MAE by $32\%$ compared to base models and by $22\%$ compared to \textit{PhAST-FE} while suffering from a $7\%$ drop in Force MAE (on average across all three GNNs).
Compared to baselines, PhAST multiplies throughput by 10-15$\times$ and divides inference time by a factor of 4-8. Compared to energy-focused PhAST enhancements, it reduces inference time by $31\%$ on average and increases throughput by a factor of $2.4\times$. These scalability gains arise both from avoiding to compute the gradient and from increasing batch size given saved memory space\footnote{The enabled increase in batch size explains how throughput and inference time improve differently: while isolated forward passes can scale linearly with batch size due to GPU parallelism (throughput), the data loading of larger batches can be a bit slower (inference time). As explained before, we keep both figures because of the theoretical/practical gains trade-off.}. Lastly, we manage to make force prediction slightly more energy-conserving by using the gradient-target loss term, although the improvement is relatively small. A more detailed analysis can be found in \cref{sec:ablation}.

\begin{table*}[h]
\centering
\begin{tabular}{lccccc}
\cmidrule[1.3pt]{1-6}
\textbf{Baseline / MAE} & E-MAE & F-MAE & EC & Throughput (s/s) & Inference time (s) \\
\cmidrule[.5pt]{1-6}
\textsl{SchNet}     & $1014$ & $70.6$ & $0$ & $519 \pm 33$ & $2050 \pm 15$ \\
\rowcolor{cycle2!5} \textsl{PhAST-FE-SchNet}     & $890$ & $\pmb{68.4}$ & $0.27$ & $2404 \pm 125$ & $593 \pm 04$ \\
\rowcolor{cycle2!9} \textsl{PhAST-SchNet}   & $\pmb{595}$ & $77.2$ & $\pmb{0.22}$ & $\pmb{6042 \pm 387}$ & $\pmb{477 \pm 03}$ \\
\cmidrule[.5pt]{1-6}
\textsl{D++}     & $913$ & $69.2$ & $0$ & $103 \pm 14$ & $10189 \pm 114$ \\
\rowcolor{cycle2!5} \textsl{PhAST-FE-D++}   & $813$ & $\pmb{67.9}$ & $0.11$ & $615 \pm 40$ & $1687 \pm 05$ \\
\rowcolor{cycle2!9} \textsl{PhAST-D++}   & $\pmb{636}$ & $83.6$ & $\pmb{0.10}$ & $\pmb{1522 \pm 205}$ & $\pmb{1259 \pm 26}$ \\
\cmidrule[.5pt]{1-6}
\rowcolor{cycle2!5} \textsl{ForceNet}     & $721$ & $68.6$ & $0.25$ & $227 \pm 25$ & $4524 \pm 47$ \\
\textsl{ForceNet-FE}     & $765$ & $190$ & $0$ & $105 \pm 14$ & $9191 \pm 18$ \\
\rowcolor{cycle2!5} \textsl{PhAST-ForceNet-FE}     & $607$ & $157$ & $0$ & $505 \pm 40$ & $2117 \pm 03$ \\
\rowcolor{cycle2!9} \textsl{PhAST-ForceNet}   & $\pmb{542}$ & $\pmb{63.9}$ & $\pmb{0.19}$ & $\pmb{1090 \pm 177}$ & $\pmb{1357 \pm 77}$ \\
\cmidrule[1.3pt]{1-6}
\end{tabular}
\caption{A comparison of energy and forces prediction, throughput and inference time of \textit{PhAST-FE} and PhAST (i.e. with force-head improvements) on the baseline GNN models on \textsl{OC20} S2EF. E-MAE and F-MAE denote respectively the Average Energy/Force MAE computed over all validation splits. Inference time (sec.) and Throughput (samples/sec. processed during inference) are averaged over 3 runs.} 
\label{tab:best-perf-s2ef}
\end{table*}

%%%%%%%%%%%%%%%%%%%%%%%%%%%%%%%%%%%%%%%%
% ABLATION
%%%%%%%%%%%%%%%%%%%%%%%%%%%%%%%%%%%%%%%%

\section{Ablation study}
\label{sec:ablation}

In this section, we provide the results of a careful ablation study assessing the contribution to both accuracy and reduction in compute time of each of the proposed components of PhAST. Note that due to the size and computational cost of GemNet and GemNet-OC, we did not conduct a full ablation study on those two models.

\begin{figure}[htb]
    \centering
    \includegraphics[width=\textwidth]{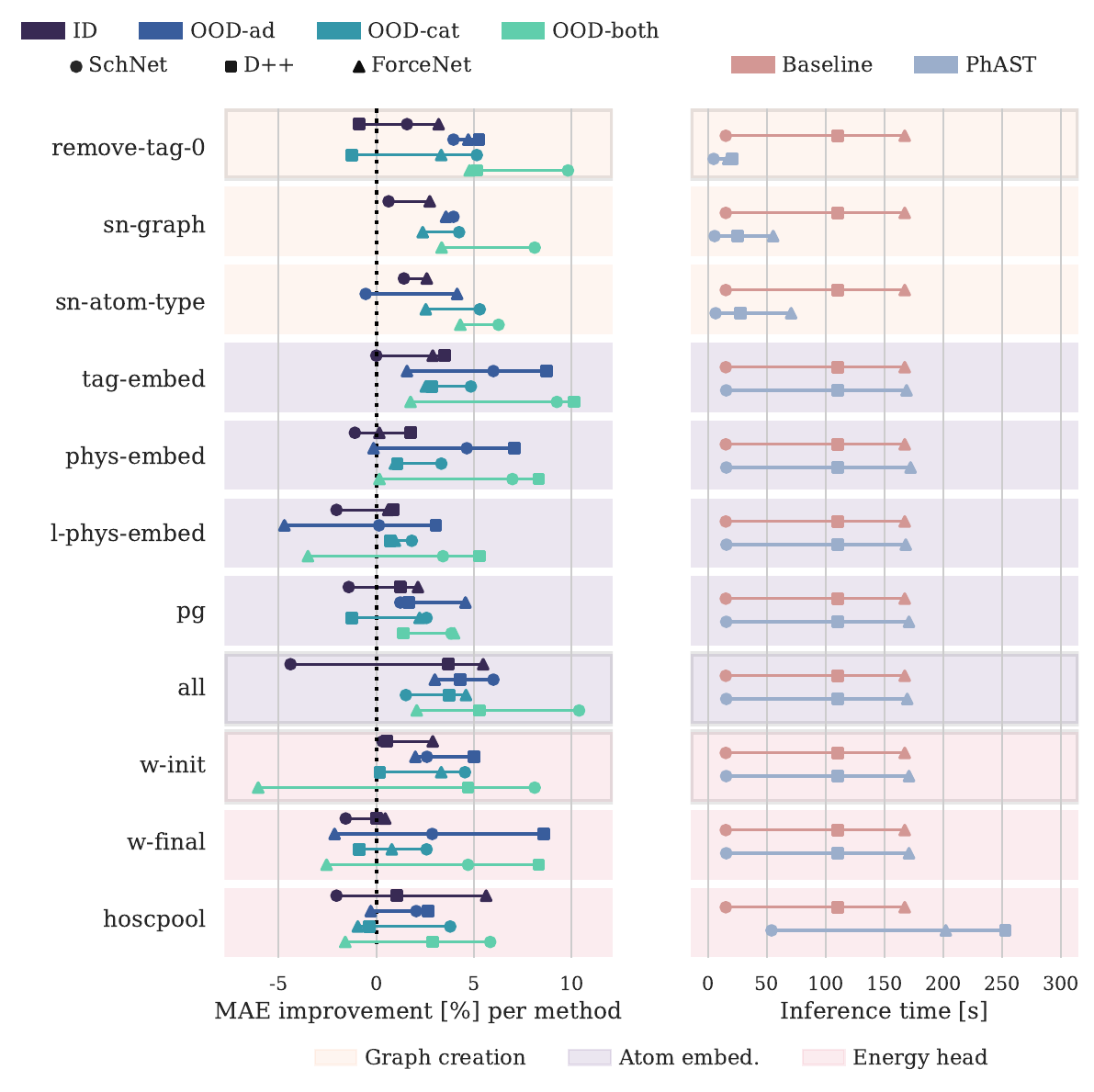}
    \caption{Ablation study results on IS2RE of our PhAST contributions about (1) the graph creation step, (2) atom embeddings and (3) the energy prediction head, all detailed in \cref{sec:method}, for SchNet, DimeNet++ and ForceNet. All changes lead to significant improvements in model performance compared to the dash line denoting the baseline (Left). All contributions have little impact on inference time except from the graph rewiring steps which divides it by several factors (Right). The best technique from each family selected for PhAST are highlighted with a darker edge around the shaded background.}
    \label{fig:ablation}
\end{figure}

\subsection{Selecting IS2RE PhAST components}
\label{subsec:ablation-is2re}

We study the accuracy and scalability gains of 

\begin{itemize}
    \item [--] the graph creation step contributions: (1) remove-tag-0 (2) one-supernode-per-graph (denoted as \textit{sn-graph} in the table) (3) one-supernode-per-atom-type (\textit{sn-atom-type}); all described in \cref{sec:graph-creation}. 
    \item [--] the atom embeddings' contributions, where we include in addition to $\mathbf{H_Z}$: (1) tag embeddings $\mathbf{H_T}$ (denoted as \textit{tag-embed} in the table) (2) physics-aware embeddings $\mathbf{H_F}$ (\textit{phys-embed}) (3) learned physics aware embeddings (\textit{l-phys-embed}) (4) period and group embeddings $\mathbf{H_P, H_G}$ (\textit{pg}) (5) all of them (\textit{all}, i.e. 1-2-4 concatenated), all described in \cref{sec:atom-embed}.
    \item [--] the energy-head contributions: (1) a weighted sum of node representation from initial encodings (denoted as \textit{w-init} in the table) (2) a weighted sum of node representation from final encodings (\textit{w-final}) (3) the hierachical pooling operator (\textit{hoscpool}), all described in \cref{sec:energy-head}.
\end{itemize}

\Cref{fig:ablation} shows graphical results of the MAE improvement computed as in Equation~\ref{eq:mae_improvement} of the ablation study whose exact numerical values can be found in \cref{app:tab:schnet}, \cref{app:tab:D++}, \cref{app:tab:forcenet} in the appendix. From \Cref{fig:ablation}, we derived the following observations:

\begin{itemize}
    \item Regarding our proposed improvements in the graph creation step, sub-surface atoms appear to contain redundant information as \textit{remove-tag-0} does not cause performance drop and aggregating it into super nodes does not yield better results. We offer two potential explanations: (1) the data generation process of DFT simulations is not optimal and tag 0 does contain redundant information (2) ML models do not manage to extract meaningful information from this repeated pattern, in which case our approach could be used by future work to demonstrate a better usage of this long range context info. \textit{remove-tag-0} also dramatically decreases compute time.  
    \item Enriched atom embeddings improve performance and generalisation, especially when adding tag information (available in the data set). Notably, the combination of all embeddings provided the best results.
    \item Regarding the energy head, the hierarchical pooling approach is not very successful, either due to the difficulty of the task or the absence of hierarchical structures; but both energy-head weighted sums are beneficial.
\end{itemize}

In conclusion, we obtain the following best components for the PhAST version of our models: remove-tag-0, full concatenation of atom embeddings (all) and predicting the system energy as a learned weighted sum of per-atom predictions, from the initial embeddings (w-init). \\

\subsection{Selecting S2EF PhAST components}

\cref{tab:best-perf-s2ef} contains the result of an ablation study comparing the options described in~\Cref{sec:force-head,subsec:s2ef} to adapt PhAST to the OC20 S2EF data set.  We study the accuracy / scalability trade-off of the following combinations:

\begin{enumerate}
    \item [--] \textit{FE}: original model, with forces predicted as the gradient of the energy prediction with respect to atom positions.
    \item [--] \textit{PhAST-FE}: PhAST enhancement of a baseline GNN model, with components selected in the IS2RE ablation study~(\Cref{subsec:ablation-is2re}).
    \item [--] \textit{PhAST-Direct}: PhAST model with the proposed direct force head.
    \item [--] \textit{PhAST-Grad}: PhAST model with direct force head and energy-grad loss from Eq.\ref{eq:energy-conserving}.
    \item [--] \textit{PhAST-Cos}: PhAST model with direct force head and cosine similarity loss from Eq.\ref{eq:cosine-similarity}.
\end{enumerate}

\begin{table*}[t!]
\centering
\begin{tabular}{lccc|cc}
\cmidrule[1.3pt]{1-6}
\textbf{Baseline / MAE} & E-MAE & F-MAE & EC & Throughput & Inference time \\
\cmidrule[.5pt]{1-6}
\textsl{SchNet-FE}                             & $1014$ & $70.6$ & $0$ &  $519 \pm 33$ & $2050 \pm 15$ \\
\rowcolor{cycle2!8} \textsl{PhAST-FE-SchNet}   & $890$ & $\pmb{68.4}$ & $0$ &  $2404 \pm 125$ & $593 \pm 04$ \\
\rowcolor{cycle2!8} \textsl{PhAST-Direct-SchNet}    & $716$ & $81.2$ & $0.27$ & $6110 \pm 390$ & $455 \pm 03$ \\
\rowcolor{cycle2!8} \textsl{PhAST-Grad-SchNet}    & $\pmb{619}$ & $79.9$ & $0.23$ & $\pmb{6164 \pm 411} $ & $456 \pm 03$ \\
\rowcolor{cycle2!8} \textsl{PhAST-Cos-SchNet}    & $667$ & $84.9$ & $-0.10$ & $6078 \pm 376 $ & $\pmb{435} \pm 02$ \\
\cmidrule[.5pt]{1-6}
\textsl{D++-FE}     & $913$ & $69.2$ & $0$ & $103 \pm 14$ & $10189 \pm 114$ \\
\rowcolor{cycle2!8} \textsl{PhAST-FE-D++}  & $813$ & $\pmb{67.9}$ & $0$ & $615 \pm 40$ & $1687 \pm 05$ \\
\rowcolor{cycle2!8} \textsl{PhAST-Direct-D++}   & $663$ & $83.7$ & $0.11$ & $1575 \pm 184$ & $1255 \pm 28$ \\
\rowcolor{cycle2!8} \textsl{PhAST-Grad-D++}    & $\pmb{659}$ & $83.7$ & $0.10$ & $\pmb{1621 \pm 180}$ & $1271 \pm 12$ \\
\rowcolor{cycle2!8} \textsl{PhAST-Cos-D++}    & $713$ & $83.9$ & $-0.19$ & $1607 \pm 171$ & $\pmb{1092 \pm 94}$ \\
\cmidrule[.5pt]{1-6} 
\textsl{ForceNet}     & $721$ & $68.6$  & $0.25$ & $227 \pm 25$ & $4524 \pm 47$ \\
\textsl{ForceNet-FE}     & $765$ & $190$ & 0 & $105 \pm 14$ & $9191 \pm 18$ \\

\rowcolor{cycle2!8} \textsl{PhAST-FE-ForceNet}   & $607$ & $157$ & $0$ & $505 \pm 40$ & $2117 \pm 03$ \\
\rowcolor{cycle2!8} \textsl{PhAST-Direct-ForceNet} & $\pmb{542}$ & $\pmb{63.9}$ & $0.19$ & $\pmb{1090 \pm 177}$ & $1357 \pm 77$ \\
\rowcolor{cycle2!8} \textsl{PhAST-Grad-ForceNet}    & $554$ & $64.0$ & $0.18$ & $1066 \pm 189$ & $1294 \pm 86$ \\
\rowcolor{cycle2!8} \textsl{PhAST-Cos-ForceNet}    & $700$ & $80.1$ & $-0.12$ & $1034 \pm 143$ & $1262 \pm 88$ \\
\cmidrule[1.3pt]{1-6}
\end{tabular}
\caption{Comparing model performance on \textsl{OC20} S2EF for baseline GNN, \textit{PhAST-FE} GNN and the different PhAST force-head proposed enhancements whose description is provided in the Baselines paragraph above. E-MAE and F-MAE denote respectively the Average Energy/Force MAE computed over all val splits. Inference time (sec.) and Throughput (samples/sec. during inference) are averaged over 3 runs.} 
\label{tab:best-perf-s2ef}
\end{table*}

From \cref{tab:best-perf-s2ef}, we draw the following observations: 

\begin{itemize}
    \item \textit{PhAST-FE} yields significant compute time and MAE improvements for all baselines, similarly to what we saw on IS2RE in ~\Cref{subsec:is2re,subsec:ablation-is2re}. This confirms its relevance and generalization capabilities. To be more concrete, \textit{PhAST-FE} leads to 13$\%$ and $4\%$ improvements in E-MAE and F-MAE, respectively, while inference time is divided by $4.2$ and throughput multiplied by $5.1\times$ (averaged over all three baselines).
    
    \item \textit{PhAST-Direct} yields additional computational gains compared to \textit{PhAST-FE}. Indeed, direct force predictions avoids computing the energy gradient, which saves memory as we do not have to compute the energy's gradient in the \textit{forward} pass. This extra memory space can be used to increase the batch size, leading to even lower total inference time. 
    
    \item\textit{PhAST-Direct} also leads to significant gains in Energy MAE, at the cost of Force MAE points. We hypothesise that this happens because SchNet and DimeNet++ are invariant models and struggle to propagate equivariant information for accurate force predictions. This hypothesis is reinforced by the fact that it is not the case for ForceNet, which processes directional information.
    
    \item Adding a new loss term to encourage energy conservation~(\textit{Grad}, \textit{Cos}) only has a small effect. Indeed, the difference between predicted forces and the energy gradient, given by the EC metric (see \cref{eq:energy-conserving}), only undergoes relatively small drops for \textit{Grad} vs the \textit{Direct} force head. However, it often leads to small performance gains, making it relevant nonetheless.
\end{itemize}

In conclusion, PhAST with direct force predictions using the gradient-target loss (\textit{Grad}) is a desirable enhancement if one targets good energy prediction and/or high scalability. However, if interested in pure molecular dynamics, we suggest using PhAST without this force head.

%%%%%%%%%%%%%%%%%%%%%%%%%%%%%%%%%%%%%%%%
% CPU
%%%%%%%%%%%%%%%%%%%%%%%%%%%%%%%%%%%%%%%%

\section{CPU Training Enhancements}
\label{cpu-train}
As described in previous sections, PhAST significantly improves the inference time of various models on the OpenCatalyst dataset. In this section, we'll also show how PhAST enables machine learning researchers to train their models on CPUs, thereby providing an additional hardware platform for different users. The greater abundance of CPUs compared to GPUs, especially in the computational chemistry community, makes the ability to effectively train models on CPUs highly desirable, as it unlocks the potential of training advanced ML models on OpenCatalyst for a larger set of users.

\begin{figure}[ht]
    \begin{subfigure}{0.48\textwidth}
    \includegraphics[width=\textwidth]{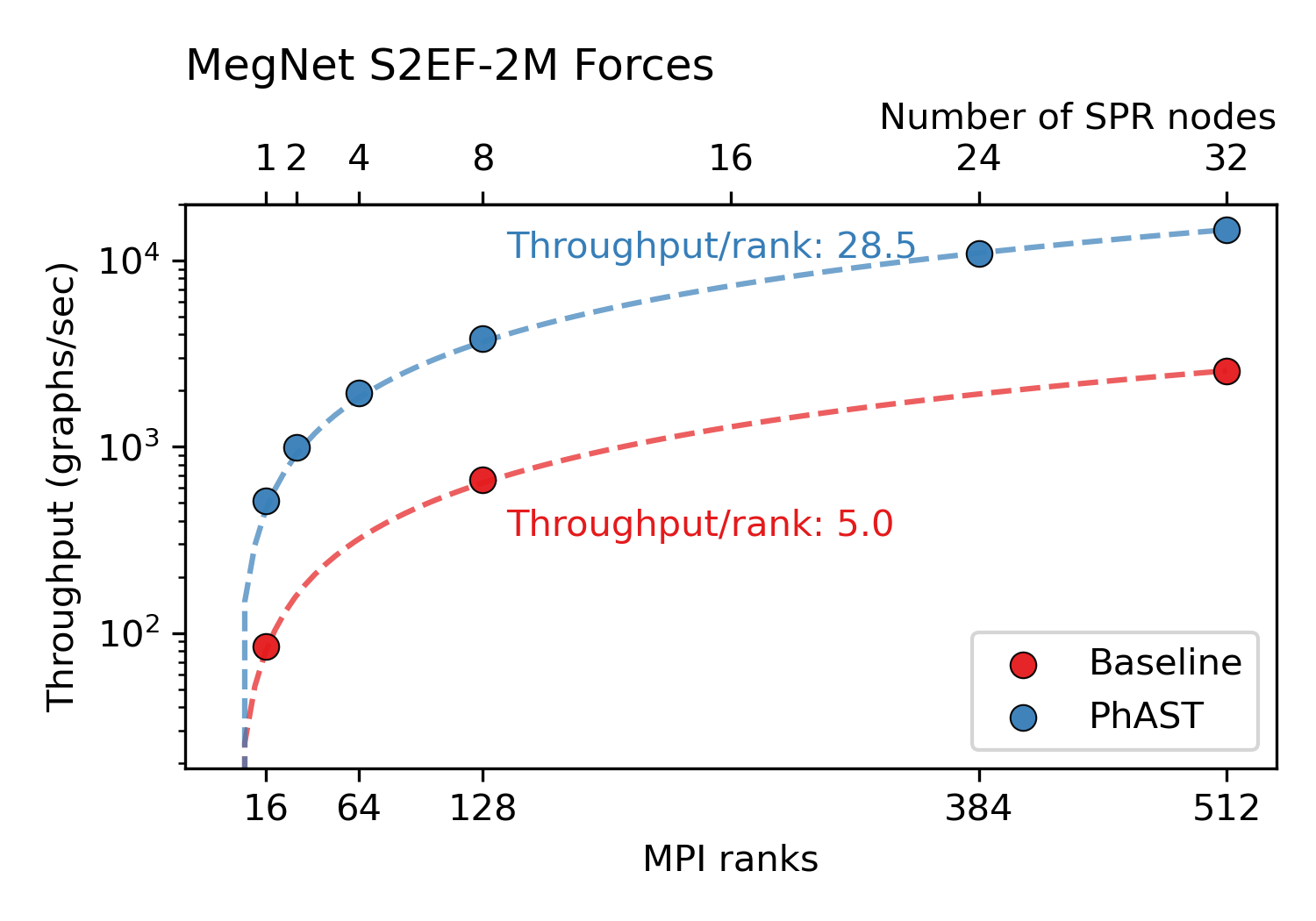}
    \subcaption{MegNet throughput on S2EF-2M on Intel Sapphire Rapids (SPR) CPU. Parallization across multiple SPR nodes (2 CPUs per node) enables significant speedback in training throughput.}
    \label{fig:s2ef-meg-throughput}
    \end{subfigure}
    % \vspace{1 cm}
    \hspace{0.02\textwidth}
    \begin{subfigure}{0.48\textwidth}
    \includegraphics[width=\textwidth]{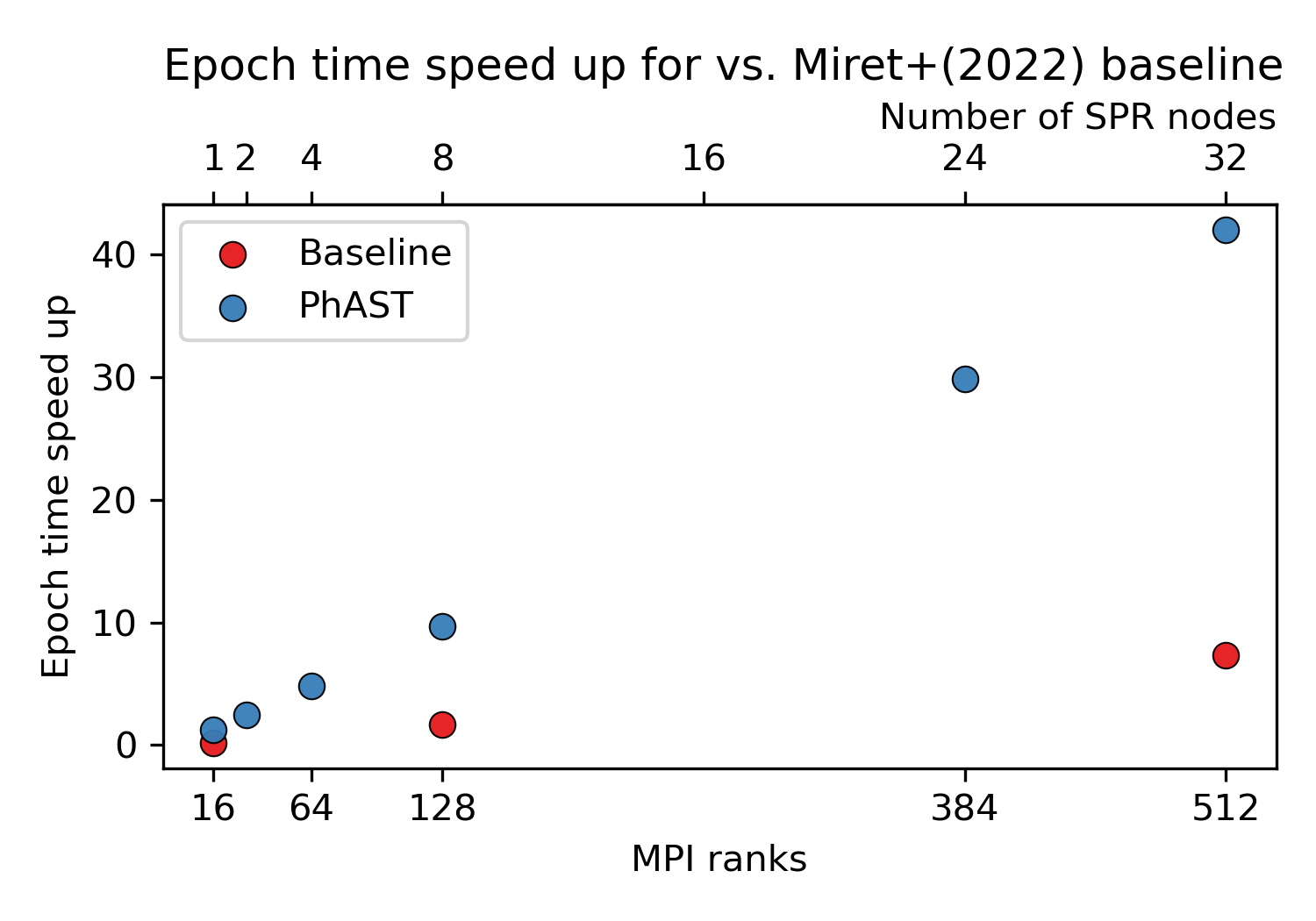}
    \subcaption{Speedup compared to \citet{miret2022open} on Intel Sapphire Rapids (SPR). At the largest degree of parallezation with 512 MPI ranks containing individual processes, we achieve ~40x speedup.}
    \label{fig:s2ef-meg-1}
    \end{subfigure}
    \captionsetup[subfigure]{skip=100pt}
    \caption{CPU-based training of MegNet \citep{chen2019graph} using 4th Gen Intel Xeon Scalable Processors known as Sapphire Rapids (SPR). The top of the x-axis specifies the number of CPU nodes with each node including 2 SPR CPUs, meaning that at the largest degree of parallization we run on 32 SPR with 64 CPUs. The bottom x-axis outlines the number of MPI ranks, which specifies the number of parallel MPI processes occuring at a given time. Using 32 SPR nodes, we can scale up to 512 MPI ranks which provide significant speedup in model trianing.}
    \label{fig:s2ef-meg-speedup}
\end{figure}

To implement PhAST CPU training, we leverage the Open MatSci ML Toolkit by \citet{miret2022open} which provides a unified platform for training deep learning on the OpenCatalyst dataset across different hardware platforms. Additionally, \citet{miret2022open} utilize the Deep Graph Library (DGL) as the platform for GNN development, which provides an additional proofpoint given that all prior experiments were performed using PyTorch Geometric. 
In this set of experiments, we focus on a distinct baseline architecture, MegNet \citep{chen2019graph}, a GNN that was constructed for chemical modeling featuring node attributes, bond attributes and graph level attributes. We performed our training on fourth generation Intel Xeon Scalable processors (named Sapphire Rapids - SPR) and investigated the scalibility and compute of PhAST across multiple CPU nodes. All such experiments perform MegNet training on S2EF-2M with hyperparameters described in \citet{miret2022open}. 

The first set of observations we can make from our study is that applying PhAST significantly increases the throughput of processed graphs for training experiments as shown in \Cref{fig:s2ef-meg-throughput}. The throughput achieved with PhAST is greater than 5$\times$ the throughput achieved without PhAST and scales more favorably up 128 MPI ranks. This confirms the general trend observed in previous experiments providing further evidence that PhAST significantly increases the compute efficiency of both inference and training on the OpenCatalyst dataset.

The second set of observations we can make relates to the training speedup achieved by PhAST using advanced CPUs seen in \Cref{fig:s2ef-meg-1}, which shows that we can achieve up 40$\times$ speedup in training time combining PhAST with advanced CPUs compared to the GPU based machines used in \citet{miret2022open}. Additionally, the increase in compute gained from PhAST is clearly shown when comparing to running the same training experiment without PhAST (shown in red on \Cref{fig:s2ef-meg-1}). At the largest degree of parallelization, regular training achieves ~5$\times$ performance gain which can attributed to more advanced hardware, which is minimal compared to the $\sim$40$\times$ speedup achieved using PhAST.

\section{Conclusion}
\label{sec:conclusion}
In this work, we presented several enhancements targeted to catalyst discovery and applicable across a variety of existing GNN models. We showed that (1) enriching atom representations with physics-based properties, (2) tailoring the graph creation to the specific task at hand, (3) weighting atoms’ importance when computing the system energy, (4) making direct force predictions with a energy conserving loss term, all reduce inference time significantly while leading to better accuracy. Besides, these gains in memory and running time make it possible to run models on CPUs, achieving up to 40$\times$ speedups and making these algorithms accessible to a greater number of researchers. Overall, our results also suggest that complex practical applications like catalyst discovery benefit from task-specific methods rather than general 3D material modeling GNNs and that performance and scalability gains can be achieved by focusing on all aspects of the pipeline instead of only the message passing block. \\

Additionally, we expect generative models to play a prominent role in catalyst discovery, replacing manual suggestion of promising new catalyst. In this paradigm, generative models require millions of calls to a GNN oracle to assess how good each catalyst is and explore the space of potential candidates accordingly. Due to its significant computational and accuracy gains, we believe that PhAST holds the potential to make a real difference, enabling the discovery of superior catalysts. This could lead to more efficient electrochemical reactions and thus contribute to reducing carbon emissions in industrial processes like fertilizer, cement, and green hydrogen production. \\

Finally, despite being designed for catalysis discovery, we anticipate that PhAST components will yield benefits in other application domains, such as QM9 \citep{ramakrishnan2014quantum}, QM7X\citep{Hoja2021qm7x} and MD17 \citep{duvenaud2015convolutional}, as well as generalizing to other GNN architectures.

\acks{This research is supported in part by ANR (French National Research Agency) under the JCJC project GraphIA (ANR-20-CE23-0009-01) as well as Samsung and Intel and was made possible thanks to Fragkiskos D. Malliaros, who also provided valuable feedback all along. Alexandre Duval acknowledges support from a Mitacs Globalink Research Award. Alex Hernandez-Garcia acknowledges the support of IVADO and the Canada First Research Excellence Fund. David Rolnick acknowledges support from the Canada CIFAR AI Chairs Program. The authors also acknowledge material support from NVIDIA and Intel in the form of computational resources, and are grateful for technical support from the Mila IDT team in maintaining the Mila Compute Cluster. The authors acknowledge the support of Kin Long Kelvin Lee in performing relevant training experiments on 4th Gen Intel Xeon Scalable Processors known as Sapphire Rapids.}

\bibliography{main}

%%%%%%%%%%%%%%%%%%%%%%%%%%%%%%%%%%%%%%%%%%%%%%%%%%%%%%%%%%%%
%%%%%%%%%%%%%%%%%%%%%%%% APPENDIX %%%%%%%%%%%%%%%%%%%%%%%%%%
%%%%%%%%%%%%%%%%%%%%%%%%%%%%%%%%%%%%%%%%%%%%%%%%%%%%%%%%%%%%

\appendix

\section{Method}

\subsection{Invariance and equivariance to symmetries}
\label{app:symmetries}

Let $\phi : V \rightarrow \R$ and $\Phi : V \rightarrow W$ be arbitrary functions where W,V are linear spaces. Let $G$ be a group describing a symmetry which we want to incorporate into $\phi$, $\Phi$ (e.g. euclidean symmetries $E(3))$. We use group representations $\rho_1: G \rightarrow GL(V)$ and $\rho_2: G \rightarrow GL(W)$, where $GL(V)$ is the space of invertible linear maps $V \rightarrow V$ to represent how the symmetries $g \in G$ are applied to vectors $X \in V, W$.
$\phi$ is an G-invariant function if it satisfies $\phi(\rho_1(g)X) = \phi(X)$, $\forall g \in G$ and $X \in V$. \\
$\Phi$ is an G-equivariant function if it satisfies $\Phi(\rho_1(g)X) = \rho_2(g) \Phi(X)$, $\forall g \in G$ and $X \in V$.

In this paper, we focus on accelerated catalysis and thus on adslab relaxed adsorption energy prediction. Like for most 3D molecular prediction tasks, we want GNNs to predict the same energy for two rotated, translated or reflected versions of the same system, since their energy is equal in real-life. Hence, we target $E(3)$-invariant models, where $E(3)$ is the Euclidean group in a 3D space (we have 3D atom positions), that is, the transformations of that 3D space that preserve the Euclidean distance between any two points (i.e. rotations, reflections, translations). Note that we do desire reflection invariance because we rotate the whole adsorbate-catalyst system and not just the adsorbate, in which case chiral molecules may have a different behaviour and shall be considered distinctly.

%%%%%%%%%%%%%%%%%%%%%%%%%%%%%%%%%%%%%%%%%%%%%%%%%%%%%%%

\subsection{Graph creation}
\label{app:sec:graph-creation}

\subsubsection{OC20}
\label{app:subsec:oc20-graph-creation}

Chanussot, Lowik, et al. \cite{chanussot2021open} create each OC20 sample by choosing a bulk material from the Materials Project database\footnote{\url{https://materialsproject.org/}}. Then, they select a surface from the bulk using Miller indices (at random) and replicate it at depth of at least $7 \textup{~\AA}$ and a width of at least $8 \textup{~\AA}$. The final slab is defined by a unit cell that is periodic in all directions with a vacuum layer of at least $20 \textup{~\AA}$ applied in the $z$ direction. Next, they pick a binding site on this surface to attach the adsorbate onto the catalyst. The graph is now a set of atoms with their 3D positions. Last but not least, edges are created between any two nodes within a cutoff distance $c=6\textup{\AA}$ of each other (considering periodic boundary conditions).

% In particular, the data set's slabs' atoms are divided into two groups: surface (tag 1) and sub-surface atoms (tag 0). The surface atoms are the closest to the adsorbate, and are allowed to move during the relaxation. Sub-surface atoms however are further away from the adsorbate, do not move during the relaxation and and were added to represent more explicitly the repeating pattern of the surface catalyst. In the following we provide further details on our graph rewiring methods:

\subsubsection{PhAST graph creation process}
\label{app:subsec:our-graph-creation}

Although well grounded, the assumptions of this graph creation process are rarely questioned. We do, with the objective of making the graph sparser and more informative for subsequent GNNs. We describe more formally the three proposals evoked in \ref{sec:graph-creation}.

\textbf{remove-tag-0}. Let $\mathcal{S} = \{i \in \mathcal{V}: t_i=0\}$ denote the set of tag 0 atoms in the atomic system. The new graph we derive has attributes $\X = \X_\mathcal{S}$ where $\X_\mathcal{S}$ is the position of all atoms except those in $\mathcal{S}$. Similarly, $\Z = \Z_\mathcal{S}$ and $\T = \T_\mathcal{S}$. The new adjacency matrix $\A_S$ 
is still defined based on cutoff distance and periodic boundary conditions: $A_{ij} = 1 \textrm{\ if\ } || \x_i - \x_j|| < c$, $0$ otherwise. But it focuses on $\X_\mathcal{S}$, thus only containing edges which do not involve atoms in $\mathcal{S}$. Same for cell offsets $\mathbf{O}$. 

\textbf{one-supernode-per-graph}. The position of the created super node is the mean of its components: $\x_s = \frac{1}{|\mathcal{S}|} \sum_{i \in \mathcal{S}} \x_i$ (with $\mathcal{S}$ as defined above). We associate it to a new characteristic number $z_s$  (corresponding to a new element in atomic table) and adjacency $A_{is} = \max(a_{ij}: j \in \mathcal{S})$. We now remove all tag-0 atoms using the \textit{remove-tag-0} method, and finally add a tag-0 attribute $t_i=0$ to the supernode. Note that we also remove self-loop for the supernode. 

\textbf{one-supernode-per-atom-type}. This extension is similar to the previous one, except that we create one supernode for each chemical element in the sub-surface catalyst. This complexify a bit the graph definition.  Let $a$ be the number of distinct elements with tag 0 in the graph ($a=1,2,3$ by construction) and $a_k$ be their characteristic number. Let $\mathcal{S}_{a_k} = \{i \in \mathcal{V} | t_i = 0 \textrm{\ and\ } z_i=a_k\}$ be the set of atoms corresponding to each distinct tag-0 element $a_k$. Each supernode $s_k$ is defined with $x_{s_k} = \frac{1}{|\mathcal{S}_{a_k}|} \sum_{i \in \mathcal{S}_{a_k}} \x_i$, $z_k=a_k$, $A_{i s_k}=\max(A_{ij}: j \in \mathcal{S}_{a_k})$ ($\forall i \in \mathcal{V}$)\textrm{\ and\ }$A_{s_{k'} s_{k}}=1$, $A_{s_{k} s_{k}}=0$. \\

For both super-node methods, we encode the number of tag-0 nodes aggregated into each super node with \textbf{Positional Encodings}~(\citep{vaswani2017attention}) to represent their "cardinal".

%%%%%%%%%%%%%%%%%%%%%%

\subsection{Atom properties for the Embedding block}
\label{app:sec:sn-type}

In atom embeddings, we use the following properties from the \texttt{mendeleev} Python package (\cite{mendeleev2014}):

\begin{enumerate}
    \item atomic radius, 
    \item atomic volume
    \item atomic density
    \item dipole polarizability
    \item electron affinity
    \item electronegativity (allen)
    \item Van-Der-Walls radius
    \item metallic radius
    \item covalent radius
    \item ionization energy (first and second order).
\end{enumerate}

%%%%%%%%%%%%%%%%%%%%%%%%%

%%%% ------------------------------------------------------

\section{Results}

\subsection{Hyperparameters}
\label{sec:ablation-hyperparameters}

\textbf{Hyperparameters}. We use each method's optimal set of parameters, provided in the config folder of the OCP repository for the IS2RE task, for the full dataset: \url{https://github.com/Open-Catalyst-Project/ocp/tree/main/configs/is2re/all}. Since ForceNet was not applied to IS2RE before, we adapted its S2EF configuration file to fit the IS2RE task. The only change is the smaller number of epochs used for DimeNet++ (10 instead of 20) and SchNet (20 instead of 30), as these additional epochs only lead to a small performance gain for a large amount of additional compute time. For PhAST models, we fine-tuned its hyperpameters to reach optimal performance. 

\subsection{GemNet IS2RE results}
\label{app:gemnets}

Results provided in~\Cref{tab:best-perf-is2re} should appear surprising: both GemNet and GemNet-OC report much better results on IS2RE than we show. This is due to the fact that there are, in general, two ways to obtain the relaxed energy: either through direct prediction as we have explained in the paper, or through \textit{relaxation}. The latter relies on an S2EF model that is iteratively applied to relax the system (positions are updated for the next step according to the current atom positions and associated predicted forces) until convergence, and the energy of the final (relaxed) structure is considered the model's output energy. This procedure is both more precise (yields better Energy MAE) but also much more computationally expensive. Both GemNet and GemNet-OC report the iterative \textit{relaxation}-based relaxed energy prediction method to evaluate the performance of the models on IS2RE, while training on the much larger S2EF data set.

In addition, as explained in~\Cref{subsec:is2re} we use the published hyper parameters. We could not afford the cost of a direct-IS2RE hyper parameter search on GemNet and GemNet-OC and therefore resulted to use their S2EF hyper parameters.

All in all, while the absolute values of Energy MAE may seem surprising, the point of~\Cref{fig:summary} and \Cref{tab:best-perf-is2re} is mainly to measure the \textit{relative} effects of PhAST on the individual models. And on this aspect, PhAST improves significantly the performance of all methods. We expect these benefits to translate to other training configurations and state-of-the-art architectures.

%%%%%%%%%%%%%%%%%%%%%%%%%%%%%%
% Throughput
%%%%%%%%%%%%%%%%%%%%%%%%%%%%%%

\subsection{Throughput results on IS2RE}
\label{app:sec:throughput}

In \cref{app:tab:throughput-is2re}, we include throughput results for IS2RE models.

\begin{table*}[h!]
\centering
%\resizebox{\textwidth}{!}
{\begin{tabular}{lccccc|c}
\cmidrule[1.3pt]{1-4}
\textbf{Baseline / MAE} & Average & Throughput (s/s) & Inference time (s) \\
\cmidrule[.5pt]{1-4}
\textsl{SchNet}   &  $683$ & $3190 \pm 302$ & $15 \pm 0.49$ \\
\rowcolor{cycle2!8} \textsl{PhAST-SchNet}    &  $\textbf{630}$  & $\pmb{16109 \pm 2000}$ &  $\pmb{5 \pm 0.36}$\\
\cmidrule[.5pt]{1-4}
\textsl{D++}    & $628$  & $191 \pm 30$  & $110 \pm 0.57$ \\
\rowcolor{cycle2!8} \textsl{PhAST-D++}    & $\textbf{595}$  & $\pmb{1021 \pm 130}$ & $\pmb{20 \pm 0.63}$ \\
\cmidrule[.5pt]{1-4}
\textsl{ForceNet}    & $654$  & $147 \pm 14$   & $167 \pm 0.96$ \\
\rowcolor{cycle2!8} \textsl{PhAST-ForceNet}  & $\textbf{616}$  & $\pmb{734 \pm 139}$ & $\pmb{33 \pm 0.60}$ \\
\cmidrule[.5pt]{1-4}
\textsl{GemNet}    & $674$  & $52 \pm 06$ & $487 \pm 0.24$ \\
\rowcolor{cycle2!8} \textsl{PhAST-GemNet}   & $\pmb{619}$  & $\pmb{306 \pm 31}$ & $\pmb{52 \pm 0.10}$ \\
\cmidrule[.5pt]{1-4}
\textsl{GemNet-OC}    & $654$  & $29 \pm 04$ & $858 \pm 1.89$ \\
\rowcolor{cycle2!8} \textsl{PhAST-GemNet-OC}   & $\textbf{616}$  & $\pmb{88 \pm 06}$ & $\pmb{257 \pm 0.09}$ \\
\cmidrule[1.3pt]{1-4}
\end{tabular}
}
\caption{Comparing model performance on \textsl{OC20} IS2RE. \textit{Average} MAE is computed over all validation splits. PhAST models all show improved performance and drastic speedups (e.g. throughput is roughly 5$\times$ higher). 
Inference times and Throughput (number of samples per second processed at inference time) are averaged over 3 runs.}
\label{app:tab:throughput-is2re}
\end{table*}

%%%%%%%%%%%%%%%%%%%%%%%%%%%%%%
% Numerical results
%%%%%%%%%%%%%%%%%%%%%%%%%%%%%%

\subsection{Numerical results of the ablation study on IS2RE}
\label{app:sec:ablation}

\textbf{Notation}. For the embedding block, \textsl{tag-embed} defines atom embeddings $\H$ using atom tag information and characteristic number: $\H = \H_Z || \H_T$, where $||$ denotes concatenation. Similarly, \textsl{phys-embed} defines $\H = \H_Z || \H_F$. \textsl{l-phys-embed} is a learnable alternative $\H = \H_Z || MLP(\H_F)$. \textsl{pg} refer to period and group embeddings: $\H = \H_Z || \H_P || \H_G$. \textsl{All} is a concatenation of all five embeddings: $\H = \H_Z || \H_T || \H_F || \H_P || \H_G$. For the graph creation step, we have defined these approaches in \ref{app:subsec:our-graph-creation} (note: \textit{sn} stands for supernode). For the energy head part, \textsl{w-init} (\textsl{w-final}) denote the weighted sum of initial (final) atom embeddings. \textsl{graclus} and \textsl{hoscpool} refer to the two hierarchical pooling approaches. 

See Tables \ref{app:tab:schnet}, \ref{app:tab:D++}, \ref{app:tab:forcenet}. The results are reported in a similar fashion as for Table 1. The symbol \textsuperscript{\Cross} indicates that a result is better than the baseline model. \textbf{Bold} font shows the best extension for each PhAST improvement category.  

%%%% ------------------------------- SCHNET

\begin{table*}[h!]
\centering
\begin{tabular}{lccccc|c}
\cmidrule[1.3pt]{1-7}
\textbf{Method / MAE} & Average & ID & OOD-ad & OOD-cat & OOD-both & Inference time (s) \\
\cmidrule[.5pt]{1-7}
\textbf{\textsl{tag-embed}}     & \textbf{648\textsuperscript{\Cross}} & \textbf{637} & \textbf{690\textsuperscript{\Cross}} & \textbf{629\textsuperscript{\Cross}} & 638\textsuperscript{\Cross} & 15.39 +/- 0.33 \\
\textsl{phys-embed}     & 662\textsuperscript{\Cross} & 644 & 700\textsuperscript{\Cross} & 639\textsuperscript{\Cross} & 654\textsuperscript{\Cross} & 15.48 +/- 0.38 \\
\textsl{l-phys-embed}     & 678\textsuperscript{\Cross} & 650 & 733\textsuperscript{\Cross} & 649\textsuperscript{\Cross} & 679\textsuperscript{\Cross} & 15.59 +/- 0.37 \\
\textsl{pg}     & 673\textsuperscript{\Cross} & 646 & 725\textsuperscript{\Cross} & 644\textsuperscript{\Cross} & 676\textsuperscript{\Cross} & 15.44 +/- 0.52 \\
\textsl{All}     & 659\textsuperscript{\Cross} & 665 & \textbf{690\textsuperscript{\Cross}} & 651\textsuperscript{\Cross} & \textbf{630\textsuperscript{\Cross}} & 15.46 +/- 0.46 \\
\cmidrule[.5pt]{1-7}
\textbf{\textsl{remove-tag-0}}     & \textbf{648\textsuperscript{\Cross}} & \textbf{627\textsuperscript{\Cross}} & \textbf{705\textsuperscript{\Cross}} & 627\textsuperscript{\Cross} & \textbf{634\textsuperscript{\Cross}} & $\pmb{4.74}$ +/- $\pmb{0.50}$ \\
\textsl{sn-graph}     & 654\textsuperscript{\Cross} & 633\textsuperscript{\Cross} & \textbf{705\textsuperscript{\Cross}} & 633\textsuperscript{\Cross} & 646\textsuperscript{\Cross} & 5.54 +/- 0.68 \\
\textsl{sn-atom-type}     & 663\textsuperscript{\Cross} & 628\textsuperscript{\Cross} & 738 & \textbf{626\textsuperscript{\Cross}} & 659\textsuperscript{\Cross} & 6.38 +/- 0.45 \\
\cmidrule[.5pt]{1-7}
\textbf{\textsl{w-init}}     & \textbf{657\textsuperscript{\Cross}} & \textbf{635\textsuperscript{\Cross}} & 715\textsuperscript{\Cross} & \textbf{631\textsuperscript{\Cross}} & \textbf{646\textsuperscript{\Cross}} & 15.37 +/- 0.48 \\
\textsl{w-final}     & 668\textsuperscript{\Cross} & 647 & \textbf{713\textsuperscript{\Cross}} & 644\textsuperscript{\Cross} & 670\textsuperscript{\Cross} & 15.41 +/- 0.38 \\
% \textsl{graclus}     & 950 & 910 & 994 & 946 & 950 & 16.65 +/- 0.60 \\
\textsl{hoscpool}     & 667\textsuperscript{\Cross} & 650 & 719\textsuperscript{\Cross} & 636\textsuperscript{\Cross} & 662\textsuperscript{\Cross} & 53.93 +/- 1.36 \\
\cmidrule[.5pt]{1-7}
\rowcolor{cycle2!8} \textsl{SchNet}     & 683 & 637 & 734 & 661 & 703 & 15.40 +/- 0.49 \\

\cmidrule[1.3pt]{1-7}
\end{tabular}
\caption{SchNet ablation study on \textsl{OC20} IS2RE. }
\label{app:tab:schnet}
\vspace{-.5em}
\end{table*}

%%%% ------------------------------- D++

\begin{table*}[h!]
\centering
\begin{tabular}{lccccc|c}
\cmidrule[1.3pt]{1-7}
\textbf{Method / MAE} & Average & ID & OOD-ad & OOD-cat & OOD-both & Inference time (s) \\
\cmidrule[.5pt]{1-7}
\textsl{\textbf{tag-embed}}     & \textbf{579\textsuperscript{\Cross}} & 551\textsuperscript{\Cross} & \textbf{659\textsuperscript{\Cross}} & 545\textsuperscript{\Cross} & \textbf{594\textsuperscript{\Cross}} & 110.04 +/- 0.89 \\
\textsl{phys-embed}     & 590\textsuperscript{\Cross} & 561\textsuperscript{\Cross} & 671\textsuperscript{\Cross} & 555 & 606\textsuperscript{\Cross} & 110.08 +/- 0.74 \\
\textsl{l-phys-embed}   & 612\textsuperscript{\Cross} & 566\textsuperscript{\Cross} & 700\textsuperscript{\Cross} & 557\textsuperscript{\Cross} & 626\textsuperscript{\Cross} & 110.12 +/- 0.75 \\
\textsl{pg}     & 624\textsuperscript{\Cross} & 564\textsuperscript{\Cross} & 710\textsuperscript{\Cross} & 568 & 652\textsuperscript{\Cross} & 110.18 +/- 0.71 \\
\textsl{All}     & 602\textsuperscript{\Cross} & \textbf{550\textsuperscript{\Cross}} & 691\textsuperscript{\Cross} & \textbf{540\textsuperscript{\Cross}} & 626\textsuperscript{\Cross} & 110.03 +/- 0.77 \\
\cmidrule[.5pt]{1-7}
\textsl{remove-tag-0}     & 610\textsuperscript{\Cross} & 576 & 684\textsuperscript{\Cross} & 568 & 627\textsuperscript{\Cross}  & $\pmb{20.15}$ +/- $\pmb{0.55}$\textsuperscript{\Cross} \\
\textsl{sn-graph}     & - & - & - & - & - & 25.04 +/- 1.54 \\
\textsl{sn-atom-type}     & - & - & - & - & - & 27.22 +/- 0.94 \\
\cmidrule[.5pt]{1-7}
\textsl{w-init}     & 611\textsuperscript{\Cross} & 568\textsuperscript{\Cross} & 686\textsuperscript{\Cross} & \textbf{560}\textsuperscript{\Cross} & 630\textsuperscript{\Cross} & 110.23 +/- 0.81 \\
\textsl{\textbf{w-final}}     & \textbf{601\textsuperscript{\Cross}} & 571 & \text{660\textsuperscript{\Cross}} & 566 & \textbf{606\textsuperscript{\Cross}} & 110.16 +/- 0.76 \\
% \textsl{graclus}     & 620\textsuperscript{\Cross} & 567\textsuperscript{\Cross} & 701\textsuperscript{\Cross} & 567 & 648\textsuperscript{\Cross} & - \\
\textsl{hoscpool}     & 618\textsuperscript{\Cross} & \textbf{565\textsuperscript{\Cross}} & 703\textsuperscript{\Cross} & 563 & 642\textsuperscript{\Cross} & 252.55 +/- 0.45 \\
\cmidrule[.5pt]{1-7}
\rowcolor{cycle2!8} \textsl{D++}  & 628 & 571 & 722 & 561 & 661 & 110.11 +/- 0.57 \\

\cmidrule[1.3pt]{1-7}
\end{tabular}
\caption{Dimenet++ ablation study on \textsl{OC20} IS2RE\textsuperscript{*} }
\label{app:tab:D++}
\small\textsuperscript{*} Unfortunately we did not manage to train DimeNet++ with the super node extensions.\\
\small For a very wide range of learning rates, the training loss consistently reached NaN values.

\end{table*}

%%%% ----------------------------- Forcenet

\begin{table*}[h!]
\centering
\begin{tabular}{lccccc|c}
\cmidrule[1.3pt]{1-7}
\textbf{Method / MAE} & Average & ID & OOD-ad & OOD-cat & OOD-both & Inference time (s) \\
\cmidrule[.5pt]{1-7}
\textsl{tag-embed}     & 640\textsuperscript{\Cross} & 639\textsuperscript{\Cross} & 690\textsuperscript{\Cross} & 616\textsuperscript{\Cross} & 617\textsuperscript{\Cross} & 168.64 +/- 0.73 \\
\textsl{phys-embed}     & 653\textsuperscript{\Cross} & 657\textsuperscript{\Cross} & 702 & 626\textsuperscript{\Cross} & 627\textsuperscript{\Cross} & 172.04 +/- 0.98 \\
\textsl{l-phys-embed}     & 667 & 654\textsuperscript{\Cross} & 734 & 626\textsuperscript{\Cross} & 650 & 167.98 +/- 0.71 \\
\textbf{\textsl{pg}}     & \textbf{634\textsuperscript{\Cross}} & 644\textsuperscript{\Cross} & \textbf{669\textsuperscript{\Cross}} & 618\textsuperscript{\Cross} & \textbf{603\textsuperscript{\Cross}} & 170.70 +/- 0.96 \\
\textsl{All}     & 637\textsuperscript{\Cross} & \textbf{622\textsuperscript{\Cross}} & 680\textsuperscript{\Cross} & \textbf{603} & 615\textsuperscript{\Cross} & 169.23 +/- 0.84 \\
\cmidrule[.5pt]{1-7}
\textbf{\textsl{remove-tag-0}}     & \textbf{628\textsuperscript{\Cross}} & \textbf{637\textsuperscript{\Cross}} & \textbf{668\textsuperscript{\Cross}} & \textbf{611\textsuperscript{\Cross}} & \textbf{598\textsuperscript{\Cross}} & $\pmb{17.06}$ +/- $\pmb{0.58}$ \\
\textsl{sn-graph}     & 635\textsuperscript{\Cross} & 640\textsuperscript{\Cross} & 676\textsuperscript{\Cross} & 617\textsuperscript{\Cross} & 607\textsuperscript{\Cross} & 55.30 +/- 1.86 \\
\textsl{sn-atom-type} & 632\textsuperscript{\Cross} & 641\textsuperscript{\Cross} & 672\textsuperscript{\Cross} & 616\textsuperscript{\Cross} & 601\textsuperscript{\Cross} & 70.55 +/- 0.15 \\
\cmidrule[.5pt]{1-7}
\textsl{w-init}      & \textbf{639\textsuperscript{\Cross}} & 639\textsuperscript{\Cross} & 687\textsuperscript{\Cross} & \textbf{611\textsuperscript{\Cross}} & 616\textsuperscript{\Cross} & 170.71 +/- 0.60 \\
\textsl{w-final}     & 660 & 655\textsuperscript{\Cross} & 716 & 627\textsuperscript{\Cross} & 644 & 170.72 +/- 1.61 \\
% \textsl{graclus}     & 671 & 622\textsuperscript{\Cross} & 722 & 634 & 646 & 172.49 +/- 1.10 \\
\textsl{hoscpool}     & 655 & \textbf{621\textsuperscript{\Cross}} & 703 & 638 & \textbf{638} & 202.03 +/- 1.23 \\
\cmidrule[.5pt]{1-7}
\rowcolor{cycle2!8} \textsl{ForceNet}     & 654 & 658 & 701 & 632 & 628 & 167.08 +/- 0.52 \\
\cmidrule[1.3pt]{1-7}
\end{tabular}
\caption{ForceNet ablation study on \textsl{OC20} IS2RE.} 
\label{app:tab:forcenet}
\vspace{-.5em}
\end{table*}

%%%%%%%%%%%%%%%%%%%%
% GR 
%%%%%%%%%%%%%%%%%%%%

\subsection{Graph-Rewiring: impact on the number of edges and nodes}
\label{app:node-count}

\begin{table}[h!]
\centering
\begin{tabular}{lcc}
\textbf{Rewiring}                    & \textbf{Atoms}      & \textbf{Edges}         \\ 
                            &            &               \\
\hline
                            &            &               \\
\textbf{Train}              &            &               \\
Full graph                  & 35 789 459 & 1 309 308 840 \\
remove-tag-0                & 32.53\%    & 16.61\%       \\
one-supernode-per-graph     & 33.81\%    & 17.76\%       \\
one-supernode-per-atom-type & 35.45\%    & 19.09\%       \\ 
                            &            &               \\
\hline
                            &            &               \\
\textbf{ID}                 &            &               \\
Full graph                  & 1 939 553  & 70 825 106    \\
remove-tag-0                & 32.54\%    & 16.65\%       \\
one-supernode-per-graph     & 33.83\%    & 17.80\%       \\
one-supernode-per-atom-type & 35.47\%    & 19.13\%       \\ 
                            &            &               \\
\hline
                            &            &               \\
\textbf{OOD-ads}            &            &               \\
Full graph                  & 1 918 704  & 69 877 652    \\
remove-tag-0                & 32.42\%    & 16.50\%       \\
one-supernode-per-graph     & 33.72\%    & 17.64\%       \\
one-supernode-per-atom-type & 35.38\%    & 18.97\%       \\
                            &            &               \\
\hline
                            &            &               \\
\textbf{OOD-cat}            &            &               \\
Full graph                  & 1 917 954  & 70 314 085    \\
remove-tag-0                & 32.88\%    & 16.78\%       \\
one-supernode-per-graph     & 34.18\%    & 17.95\%       \\
one-supernode-per-atom-type & 35.90\%    & 19.34\%       \\
                            &            &               \\
\hline
                            &            &               \\
\textbf{OOD-both}           &            &               \\
Full graph                  & 2 094 709  & 80 074 123    \\
remove-tag-0                & 31.12\%    & 15.33\%       \\
one-supernode-per-graph     & 32.31\%    & 16.37\%       \\
one-supernode-per-atom-type & 34.17\%    & 17.83\%       \\
                            &            &               \\
                            &            &               \\
\end{tabular}
\caption{Comparison of the number of nodes and edges in the original 5 datasets (training and 4 validation splits) and the remaining number of nodes and edges after the various rewiring strategies are performed. We can see that our rewiring methods generally remove 65+\% of the atoms and 80+\% of the edges.}
\end{table}

\end{document}